\documentclass[]{bytedance_seed}

\usepackage[toc,page,header]{appendix}

\usepackage{minitoc}
\usepackage{amssymb}
\usepackage{multirow}
\usepackage{makecell}

\newcommand{\postln}{Post-LN}
\newcommand{\preln}{Pre-LN}
\newcommand{\ours}{\textsc{Keel}}
\newcommand{\model}{KEEL}

\title{Post-LayerNorm Is Back: Stable, ExpressivE, and Deep}

\author[*]{Chen Chen}
\author[*,\dagger]{\ \ Lai Wei}

\affiliation{ByteDance Seed}

\contribution[*]{Equal Contribution}
\contribution[\dagger]{Corresponding authors}

\abstract{
Large language model (LLM) scaling is hitting a wall. Widening models yields diminishing returns, and extending context length does not improve fundamental expressivity. 
In contrast, depth scaling offers theoretically superior expressivity, yet current Transformer architectures struggle to train reliably at extreme depths.
We revisit the Post-LayerNorm (Post-LN) formulation, whose instability at scale caused its replacement by Pre-LN in modern LLMs. We show that the central failure mode of Post-LN arises from the ResNet-style residual pathway, which introduces gradient vanishing in deep networks.
We present \ours{}, a Post-LN Transformer that \textbf{replaces this residual path with a Highway-style connection}. This modification preserves the gradient flow through the residual branch, preventing signal vanishing from the top layers to the bottom. Unlike prior methods, \ours{} enables stable training at extreme depths without requiring specialized initialization or complex optimization tricks.
\textsc{Keel} trains robustly at depths \textbf{exceeding 1000 layers} and consistently improves perplexity and depth-scaling characteristics over Pre-LN. These findings indicate that Post-LN, when paired with a Highway-style connection, provides a simple and effective foundation for building deeply scalable LLMs, opening the possibility for future infinite-depth architectures.
}

\date{\today}
\correspondence{Lai Wei at \email{laiwei.future@bytedance.com}}

\begin{document}
\maketitle

\begin{figure}[htbp]
    \centering
    \vspace{-20pt}
    \begin{subfigure}[b]{0.325\linewidth}
        \centering
        \includegraphics[width=\linewidth]{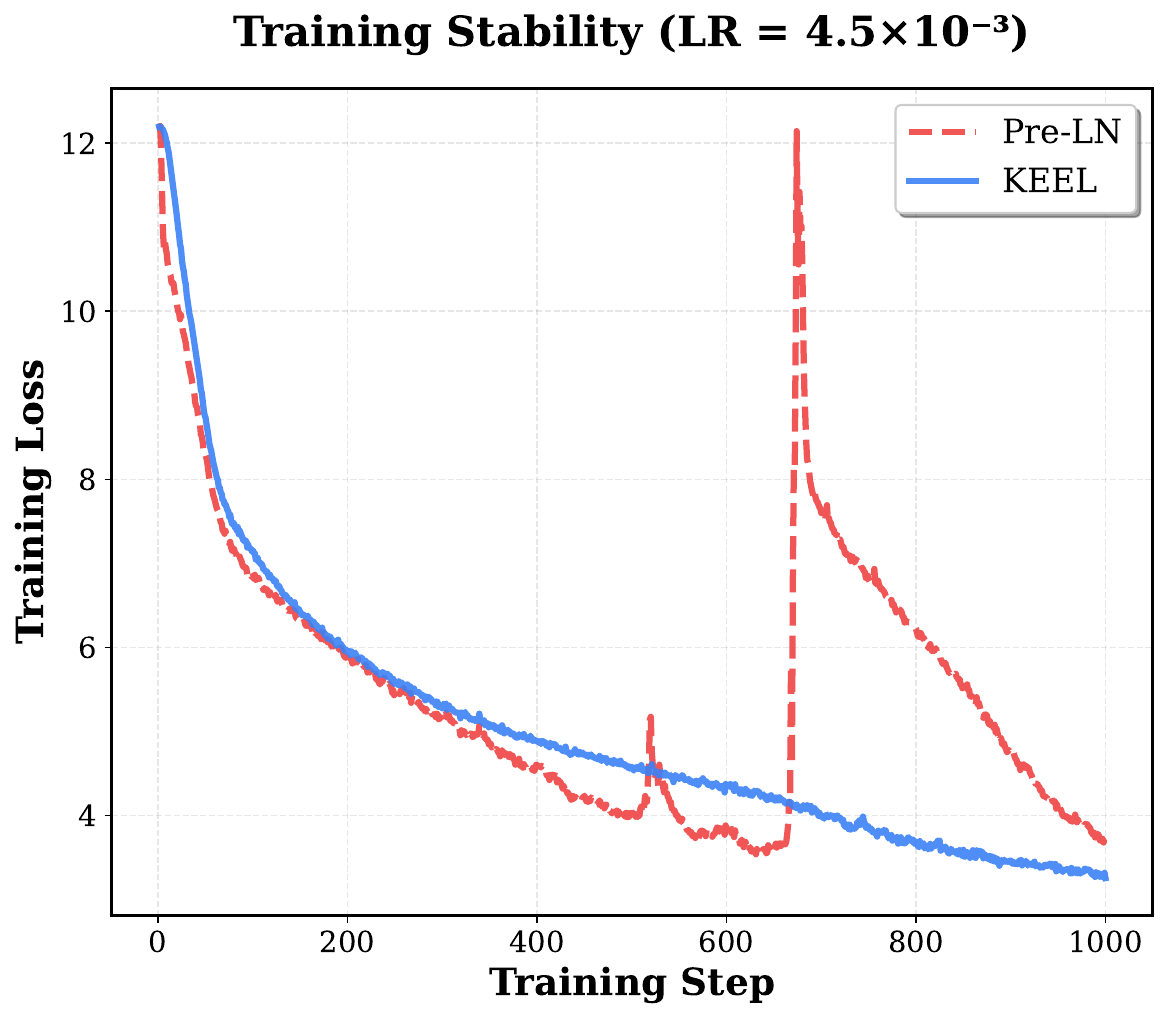}
        \caption{Training Stability}
        \label{fig:stable}
    \end{subfigure}
    \begin{subfigure}[b]{0.325\linewidth}
        \centering
        \includegraphics[width=\linewidth]{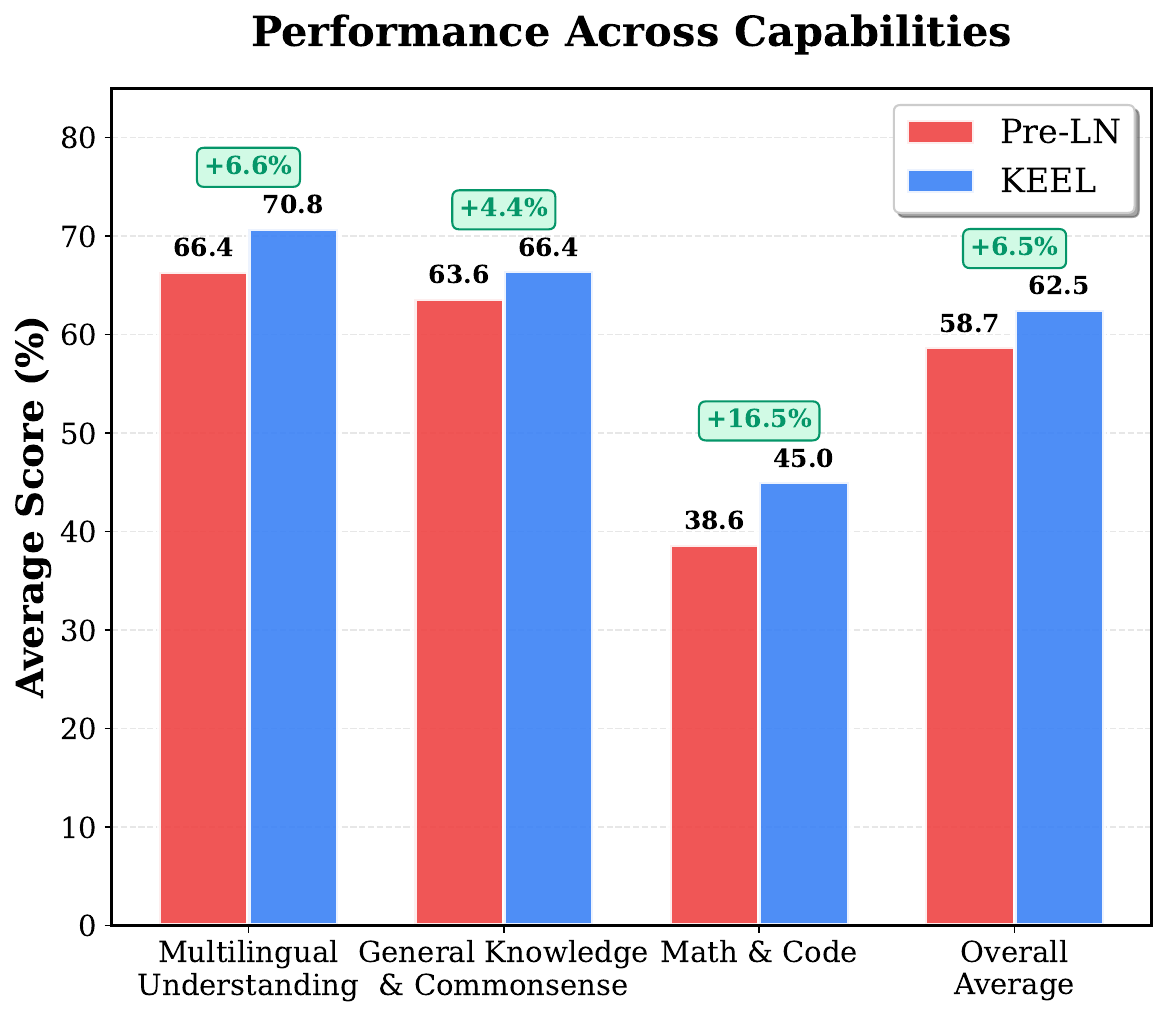}
        \caption{Expressiveness}
        \label{fig:expressive}
    \end{subfigure}
    \begin{subfigure}[b]{0.325\linewidth}
        \centering
        \includegraphics[width=\linewidth]{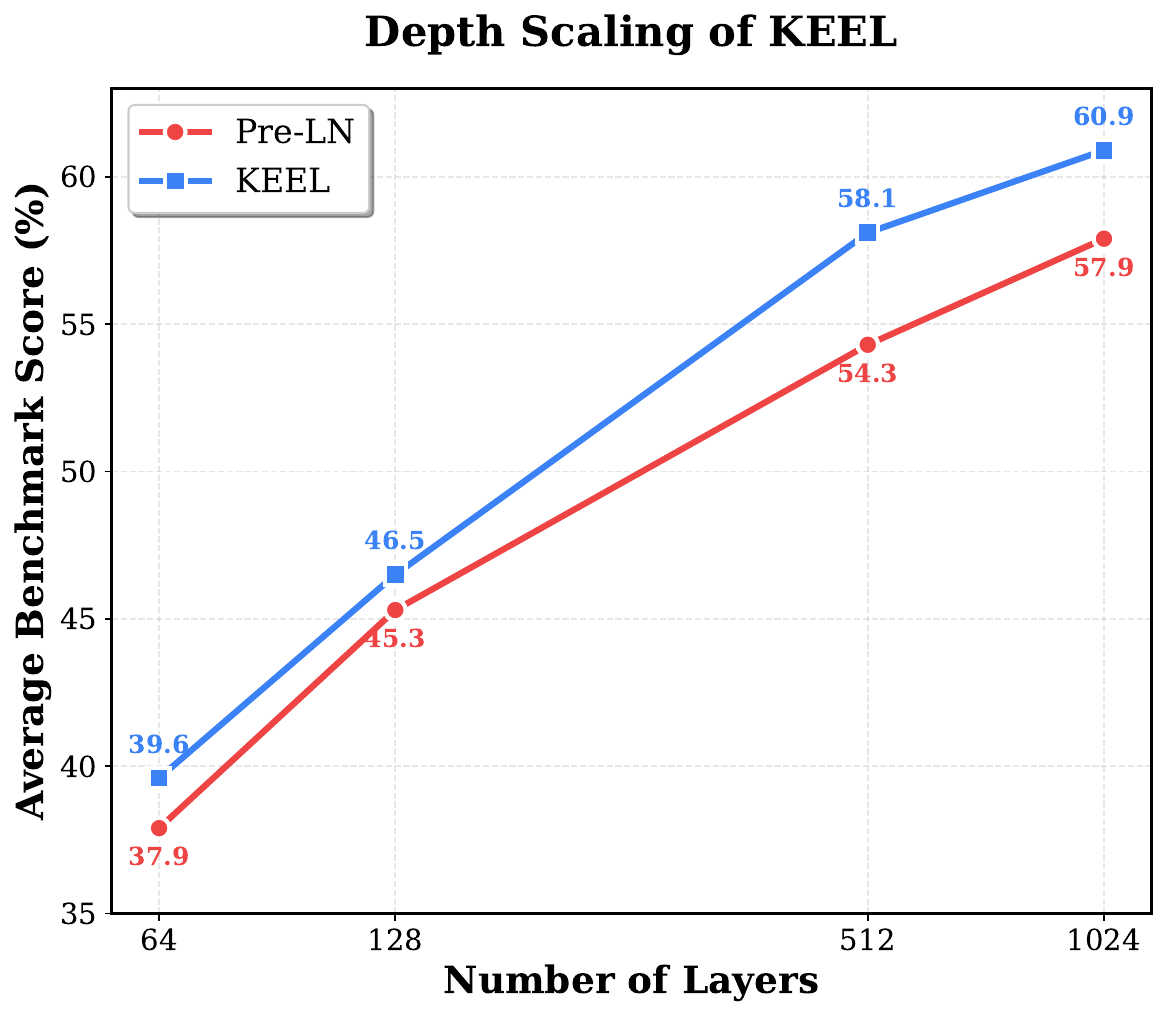}
        \caption{Depth Scaling}
        \label{fig:deep}
    \end{subfigure}
    \caption{\textbf{\model{} enables stable, expressive, and deep LLM training.} 
    (a) \textbf{Training Stability:} \ours{} maintains smooth convergence at aggressive learning rates, while Pre-LN exhibits severe instability under the same configuration. 
    (b) \textbf{Expressiveness:} \ours{} demonstrates superior performance across all capability domains, particularly in Math \& Code (+16.5\%). 
    (c) \textbf{Depth Scaling:} \ours{} consistently outperforms Pre-LN across all depths (64-1024 layers). 
    Together, these results demonstrate that \ours{}'s architectural improvements enable stable optimization of ultra-deep networks with enhanced learning efficiency and model expressiveness.
    }
    \label{fig:keel_overview}
    \vspace{-20pt}
\end{figure}

\newpage

\section{Introduction}

Large language model (LLM) progress has been driven primarily by scaling: bigger models, longer context windows, and larger training corpora. Yet these conventional scaling axes are beginning to show diminishing returns. Width scaling saturates quickly, context scaling grows increasingly expensive, and parameter growth alone does not unlock qualitatively new behaviors. As a result, LLM scaling is hitting a wall, and there is increasing interest in architectural directions that can deliver more expressivity per parameter.

Depth scaling offers a promising path forward. In principle, deeper networks can represent exponentially richer functions and support more hierarchical reasoning. However, current LLM architectures struggle to capitalize on depth. Training becomes increasingly unstable at extreme depths, and even when optimization succeeds, depth scaling delivers substantially worse returns than width scaling under current architectures.

The placement of Layer Normalization (LN), which is a seemingly simple architectural choice, has an enormous effect on depth scaling. The original Transformer architecture used Post-LayerNorm (\postln{}), but modern LLMs overwhelmingly adopt Pre-LayerNorm (\preln{})~\citep{gpt3, llama}. \preln{} stabilizes early training by normalizing each sublayer’s input, preventing the divergence commonly seen in deep \postln{} networks~\citep{xiong2020layernorm}. However, \preln{} introduces its own structural limitations: it weakens gradient propagation and reduces the effective contribution of deeper layers~\citep{nguyen2019transformers}. As models grow deeper, this results in poor depth scaling and representational expressivity, limiting the potential of depth as a new scaling axis.

In contrast, \postln{} maintains large gradient signals in deeper layers, which can support superior depth scaling. Yet its training instability has made it unsuitable for LLM-scale models. When the residual output and transformed features are summed and then normalized, gradients in LayerNorm can exhibit extreme variability, especially in deep regimes~\citep{xiong2020layernorm}. Previous attempts to revive \postln{}, such as DeepNorm~\citep{deepnorm}, Admin~\citep{admin}, and more recent hybrid normalization strategies~\citep{mixln, hybridnorm}, mitigate some failure modes but do not fundamentally resolve the gradient pathologies of \postln{} LLMs, nor do they demonstrate reliable behavior at the depths needed to break the scaling limits of today.

To understand the root cause of these instabilities, we formally analyze the gradient dynamics of \postln{}. We derive bounds on the backward signal and show that the ResNet-style residual path is the primary source of gradient vanishing. These issues arise not from normalization itself, but from the way that residual and transformed activations are mixed before normalization.

Motivated by this analysis, we consider a small yet impactful architectural change: replace the ResNet-style residual branch with a simplified Highway-style connection, and re-express its gradient dynamics under the same theoretical framework. Our results show that this Highway-style pathway provides provable control of gradient magnitudes, allowing signals to propagate through depth without vanishing. Crucially, it maintains the inter-layer coupling that makes Post-LN expressive, while suppressing the unstable mixing that previously made Post-LN difficult to train.

Guided by these findings, we introduce \ours{}, a Post-LN architecture that incorporates a lightweight Highway-style gated connection~\citep{highway}. The gate dynamically balances carry and transform signals, regulating both forward and backward information flow. This simple modification stabilizes Post-LN at scale, enabling it to realize its expressivity advantages without special initialization or customized residual scaling.

Empirically, \ours{} delivers substantial gains in depth scalability and model performance, effectively addressing the training stability issues often associated with traditional deep architectures. By integrating a Highway-style pathway with a revived \postln{} configuration, \ours{} enables robust training at depths exceeding 1000 layers.\footnote{In this paper, unless otherwise specified, ``layer'' refers to the total count of residual connections, which includes both the Attention and Feed-Forward Network (FFN) layers.} While standard \postln{} or \preln{} architectures often exhibit severe instability when subjected to aggressive learning rates, \ours{} maintains smooth convergence, suggesting a more well-conditioned optimization landscape.

This newfound stability does not come at the cost of representational power. Instead, \ours{} demonstrates superior expressiveness across the entire depth spectrum ranging from 64 to 1024 layers. These gains are particularly pronounced in specialized capability domains such as Math and Code, where the model achieves a +16.5\% performance increase over Pre-LN baselines. Collectively, these results suggest that the architectural improvements in \ours{} facilitate a more efficient learning process, allowing for the stable optimization of ultra-deep networks. By breaking the conventional depth-scaling barrier, this approach establishes a practical and powerful framework for the next generation of large language model scaling.
\section{Preliminary}
\label{sec:prelim}

In this section, we review the architectural components relevant to depth scaling and gradient propagation in deep learning. We revisit Highway Networks and Residual Networks, and then summarize the Post-LN and Pre-LN normalization schemes that define the operational behavior of modern Transformer blocks.

\subsection{Highway Networks}

Highway Networks~\cite{highway} were introduced as an early mechanism for training very deep feed-forward architectures. Their key idea is to allow hidden layers to adaptively regulate how much of their input is transformed versus directly carried forward.  
Given an input $\mathbf{x}$ and transformation $\mathbf{F}(\mathbf{x})$, a Highway layer computes:
\begin{equation}
\mathbf{x}_{l+1} 
= \mathbf{T}(\mathbf{x}_{l}) \odot \mathbf{F}(\mathbf{x}_{l}) 
 + \mathbf{C}(\mathbf{x}_{l}) \odot \mathbf{x}_{l},
\end{equation}
where $\mathbf{T}(\mathbf{x}_{l})$ is a typically trainable gate, $\mathbf{C}(\mathbf{x}_{l})$ is typically set to $1-\mathbf{T}(\mathbf{x}_{l})$ and $\odot$ denotes elementwise multiplication.
This gating mechanism ensures that gradients can bypass transformations when necessary, preventing gradient attenuation.

\subsection{Residual Networks}

Residual Networks~\cite{resnet} replace the Highway gate with a fixed identity path, computing:
\begin{equation}
\mathbf{x}_{l+1} = \mathbf{x}_{l} + \mathbf{F}(\mathbf{x}_{l}).
\end{equation}

This unconditional skip connection was later adopted in Transformers for both MHA and FFN sublayers.  
However, the interaction between this residual path and normalization plays a decisive role in stability. Specifically, whether LayerNorm is applied before or after the residual addition determines how gradients propagate and how activations accumulate across depth.

\subsection{Post-LN: The Original Transformer Formulation}

The original Transformer architecture~\citep{transformer} applies LayerNorm after the residual addition:
\begin{equation}
\mathbf{x}_{l+1} 
= \text{LN}\!\left(\mathbf{x}_{l} + \mathbf{F}(\mathbf{x}_{l})\right).
\end{equation}

Because the residual and transformed signals are jointly normalized, this design induces strong forward and backward coupling across layers. Prior work~\citep{xiong2020layernorm} has shown, however, that this coupling can lead to optimization difficulty at scale.  
During backpropagation, gradients must pass through the Jacobian of LayerNorm applied to a sum of two potentially misaligned signals:
\begin{equation}
\frac{\partial \mathbf{x}_{l+1}}{\partial \mathbf{x}_{l}}
= J_{\text{LN}}\bigl(\mathbf{x}_{l} + \mathbf{F}(\mathbf{x}_{l})\bigr),
\end{equation}
making the gradient sensitive to activation statistics and susceptible to attenuation in deep networks.

To mitigate this issue, DeepNorm~\cite{deepnorm} introduces a depth-dependent scaling factor on the residual branch:
\begin{equation}
\mathbf{x}_{l+1} = 
\text{LN}\!\left(\alpha\,\mathbf{x}_{l} + \mathbf{F}(\mathbf{x}_{l})\right),
\end{equation}
where $\alpha$ is chosen according to theoretical prescriptions. Besides, it down-scales the weight initialization by a factor $\beta$. For the decoder-only architectures, $\alpha$ and $\beta$ are set to $L^{0.25}$ and $L^{-0.25}$, respectively.

\subsection{Pre-LN: The Modern Standard in LLMs}

The Pre-LN formulation~\citep{xiong2020layernorm} inverts the normalization placement:
\begin{equation}
\mathbf{x}_{l+1} 
= \mathbf{x}_{l} + \mathbf{F}(\text{LN}(\mathbf{x}_{l})).
\end{equation}

Here, the residual path bypasses normalization entirely, yielding a clean identity gradient path:
\begin{equation}
\frac{\partial \mathbf{x}_{l+1}}{\partial \mathbf{x}_{l}} 
= \mathbf{I} + \frac{\partial \mathbf{F}}{\partial \mathbf{x}_{l}}.
\end{equation}

This enables stable optimization without special initialization and has become the default in modern LLMs such as GPT-3, PaLM, and LLaMA.  
However, because gradients flow predominantly through the identity connection, deeper layers often contribute diminishingly to the update signal, reducing depth utilization and harming scaling behavior~\citep{sun2025curse}.

\subsection{Hybrid Variants: Interpolating Between Post-LN and Pre-LN}

Several recent approaches~\cite{hybridnorm, mixln} explore hybrid normalization placements that attempt to combine the advantages of Post-LN and Pre-LN. Two representative strategies are:
\begin{itemize}
\item HybridNorm~\cite{hybridnorm}, which interleaves Post-LN and Pre-LN blocks throughout the network to blend their respective optimization and expressivity properties,
\item Mix-LN~\cite{mixln}, which applies Post-LN in lower layers and transitions to Pre-LN in upper layers, aiming to leverage stronger representational coupling at the bottom while retaining stability in deeper regions.
\end{itemize}

These hybrid designs provide improved robustness over pure Post-LN and can outperform pure Pre-LN in certain regimes. However, they do not fundamentally resolve the gradient degeneration inherent to Post-LN at very large depths, nor do they provide principled guarantees for stable scaling in the extreme-depth setting required for future LLM architecture development.

\section{\model{}}

We introduce \ours{}, a novel architecture designed to stabilize training in deep LLMs. The forward propagation for the $l$-th layer is defined as:

\begin{equation}
    \mathbf{x}_{l+1} = \operatorname{LN}\left(\textcolor{red}{\alpha} \mathbf{x}_{l} + \mathcal{F}_{l}(\textcolor{red}{\operatorname{LN}}( \mathbf{x}_{l}))\right)
\end{equation}

Compared to the vanilla \postln{} architecture, our method incorporates two critical structural modifications.

\textbf{Highway-style Residual Scaling:} We introduce a scalar $\alpha$ to weight the skip connection, creating a highway-like structure. Based on our gradient flow analysis (detailed in Section~\ref{sec:ana_keel}), we set $\textcolor{red}{\alpha = L}$, where $L$ represents the total number of sub-layers (including both Attention and FFN layers)\footnote{Setting $\alpha = L$ is critical for maintaining training stability in very large-scale or deep models. For smaller architectures where vanishing or exploding gradients are less pronounced, $\alpha$ can be treated as a tunable hyperparameter ($\alpha > 1$) to potentially accelerate convergence.}. Unlike standard gating mechanisms that require coefficients to sum to 1 (e.g., $(1-\lambda)\mathbf{x} + \lambda \mathcal{F}(\mathbf{x})$), we rely on the final \postln{} to normalize the output magnitude, rendering explicit variance constraints on the summation unnecessary.
    
\textbf{Residual Branch Normalization:} We inject an additional Layer Normalization step into the input of the residual function $\mathcal{F}_l$. While this conceptually aligns the residual branch with a \preln{} formulation (where the input is normalized), the global architecture retains the \postln{} topology. As discussed in Section~\ref{sec:discussion}, this additional normalization is crucial for stabilizing the gradient flow through the residual branch, preventing the attenuation often seen in deep networks.

\subsection{Implementation Details}

To ensure optimal performance and stability, we adopt the following implementation strategies for \ours{}.

\begin{figure}[t]
    \centering
    \includegraphics[width=0.75\linewidth]{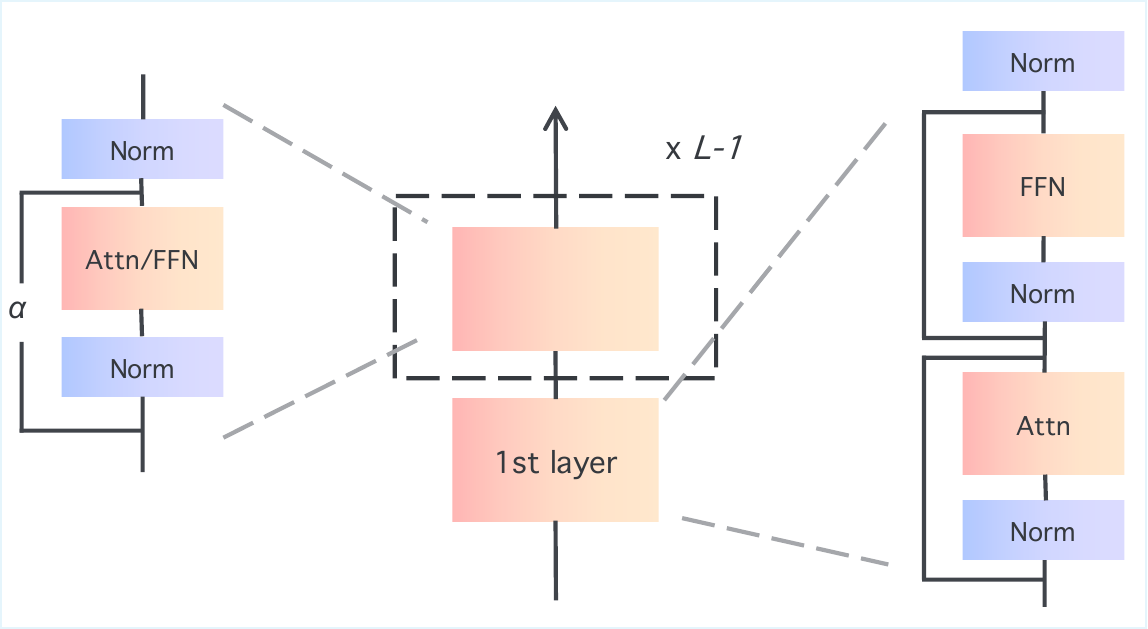}
    \caption{Illustration of our \ours{} architecture.}
    \label{fig:keel_method}
\end{figure}

\textbf{Input Layer Initialization:} As shown in Figure~\ref{fig:keel_method}, we remove the \postln{} for the very first attention layer as well as the scaling factor $\alpha$ for both the very first attention and FFN layers. Consequently, they effectively degrade to standard \preln{} or \postln{} blocks, ensuring stable signal initialization from the embedding layer.

\textbf{Hyperparameters:} Empirically, \ours{} allows for and benefits from a larger learning rate compared to standard \preln{} baselines, accelerating convergence.
    
\textbf{Normalization Configuration:} All Layer Normalization operations utilize learnable affine weights ($\boldsymbol{\gamma}$) but omit the additive bias term ($\boldsymbol{\beta} = 0$) to improve parameter efficiency and stability.

\subsection{Instability of \postln{} LLMs: A Gradient Perspective}
\label{sec:ana_post}

We first define the operation of the RMS-based Layer Normalization (LN) used in the $l$-th layer. Given an input vector $\mathbf{x}$, the normalization is formulated as:
\begin{equation}
    \operatorname{LN}(\mathbf{x}) = \frac{\mathbf{x}}{\|\mathbf{x}\|_2} \odot \boldsymbol{\gamma},
\end{equation}
where $\odot$ denotes element-wise multiplication, and $\boldsymbol{\gamma} \in \mathbb{R}^d$ is a learnable affine transformation parameter. We define the scalar magnitude of this parameter as $\gamma = \|\boldsymbol{\gamma}\|_\infty$.

In a standard \postln{} architecture, the forward propagation of the $l$-th and $(l+1)$-th sub-layers is formulated as:
\begin{align}
    \mathbf{x}_{l+1} &= \operatorname{LN}_{l}\left(\mathbf{x}_{l} + \mathcal{F}_l(\mathbf{x}_{l})\right), \\
    \mathbf{x}_{l} &= \operatorname{LN}_{l-1}\left(\mathbf{x}_{l-1} + \mathcal{F}_{l-1}( \mathbf{x}_{l-1})\right).
\end{align}
Here, the residual connection is a simple summation. Let $\mathbf{z}_l = \mathbf{x}_{l} + \mathcal{F}_l(\mathbf{x}_{l})$ denote the pre-normalization state.

During backpropagation, the gradient flows from the loss $\mathcal{L}$ through the layers via the chain rule:
\begin{equation}
    \frac{\partial \mathcal{L}}{\partial \mathbf{x}_{l}} 
    = \frac{\partial \mathcal{L}}{\partial \mathbf{x}_{l+1}} \frac{\partial \mathbf{x}_{l+1}}{\partial \mathbf{x}_{l}}.
\end{equation}
Expanding the Jacobian $\frac{\partial \mathbf{x}_{l+1}}{\partial \mathbf{x}_{l}}$ yields:
\begin{equation}
    \frac{\partial \mathbf{x}_{l+1}}{\partial \mathbf{x}_{l}} = 
    \frac{\partial \operatorname{LN}_{l}(\mathbf{z}_l) }{\partial \mathbf{z}_l} 
    \frac{\partial \mathbf{z}_l}{\partial \mathbf{x}_l} = 
    \frac{\partial \operatorname{LN}_{l}(\mathbf{z}_l)}{\partial \mathbf{z}_l} \left(\mathbf{I} + \frac{\partial \mathcal{F}_l}{\partial \mathbf{x}_{l}}\right).
\end{equation}
Focusing on the gradient flow through the residual connection, the Jacobian magnitude is determined by the normalization step. Consequently, the gradient magnitude satisfies:
\begin{equation}
    \mathbf{J}^*_{\text{LN}_{l}}(\mathbf{z}_l) = \left\| \frac{\partial \operatorname{LN}_{l}(\mathbf{z}_l)}{\partial \mathbf{z}_l} \right\|_2 
    = \mathcal{O}\left( \frac{\gamma_{l}}{\sqrt{2}\gamma_{l-1}} \right).
\end{equation}
Typically, $\gamma$ is initialized to 1. The cumulative gradient magnitude across $L$ layers thus scales as:
\begin{equation}
    \prod_{l=1}^{L} \mathbf{J}^*_{\text{LN}_{l,1}}(\mathbf{z}_l) = \mathcal{O}\left(\frac{1}{2^{\frac{L}{2}}}\right).
\end{equation}
This indicates that in standard \postln{}, the gradient signal exponentially decays as it propagates to lower layers, hindering the training of deep models.

\subsection{\model{}: Stabilizing Deep LLMs}
\label{sec:ana_keel}

To mitigate this instability, \ours{} introduces a scaling factor $\alpha$ and an additional normalization step. The forward pass is reformulated as:
\begin{align}
    \mathbf{x}_{l+1} &= \operatorname{LN}_{l,1}\left(\alpha \mathbf{x}_{l} + \mathcal{F}_l(\operatorname{LN}_{l,2}( \mathbf{x}_{l}))\right), \\
    \mathbf{x}_{l} &= \operatorname{LN}_{l-1,1}\left(\alpha \mathbf{x}_{l-1} + \mathcal{F}_{l-1}(\operatorname{LN}_{l-1,2}( \mathbf{x}_{l-1}))\right).
\end{align}
With the reintroduction of $\alpha$ to scale the shortcut branch, the denominator in the gradient derivation changes. The gradient magnitude through the $l$-th sub-layer's residual connection satisfies:
\begin{equation}
    \mathbf{J}^*_{\text{LN}_{l,1}}(\mathbf{z}_l) = \mathcal{O}\left( \frac{\gamma_{l,1} \alpha}{\sqrt{\gamma_{l-1,1}^2\alpha^2 + \gamma_{l,2}^2}} \right).
\end{equation}
We propose setting $\alpha = L$. Analyzing the asymptotic behavior as $L \rightarrow \infty$, we observe:
\begin{equation}
    \lim_{L \rightarrow \infty} \prod_{l=1}^{L} \mathbf{J}^*_{\text{LN}_{l,1}}(\mathbf{z}_l) 
    = \lim_{L \rightarrow \infty} \left( \frac{\alpha}{\sqrt{\alpha^2 + 1}} \right)^L 
    = \lim_{L \rightarrow \infty} \left[ \left(1 + \frac{1}{L^2}\right)^{-\frac{L}{2}} \right] 
    = 1.
\end{equation}


This limit confirms that our formulation prevents gradient vanishing. 

\subsection{\model{} is a \postln{} Architecture}

We classify \ours{} as a \postln{} architecture. Although \ours{} includes a normalization step inside the transformation branch $\mathcal{F}$ (similar to \preln{}), its structure is fundamentally \postln{}. This is because the distinction between \preln{} and \postln{} depends on the shortcut branch, not the input to the transformation function.

First, notice that in deep networks, the input to $\mathcal{F}$ is effectively normalized in both architectures.

\textbf{In Pre-LN:} We explicitly normalize the input to the function: $\mathcal{F}(\operatorname{LN}(\mathbf{x}_l))$.

\textbf{In Post-LN:} The input $\mathbf{x}_l$ comes directly from the previous layer's LayerNorm. Therefore, the function $\mathcal{F}(\mathbf{x}_l)$ already receives normalized input.

More specifically, If we view the $\mathbf{x}_l$ as the input of the LayerNorm from the previous layer, we can write the standard Post-LN forward pass as:
\begin{equation}
\mathbf{x}_{l+1}^{\text{Post}} = \operatorname{LN}_{l-1}(\mathbf{x}_l) + \mathcal{F}_l(\operatorname{LN}_{l-1}(\mathbf{x}_l)))
\end{equation}

Comparing this to \preln{}:
\begin{equation}
\mathbf{x}_{l+1}^{\text{Pre}} = \mathbf{x}_l + \mathcal{F}_l(\operatorname{LN}_l(\mathbf{x}_l))
\end{equation}

The input to the transformation branch $\mathcal{F}$ is normalized in both cases. \textbf{The real difference is the shortcut.} In \preln{}, the shortcut preserves the unnormalized scale of the previous layers, whereas in \postln{}, the shortcut itself is normalized.

Similarly, \ours{} can be written as:
\begin{equation}
\mathbf{x}_{l+1}^{\text{Keel}} = \alpha \operatorname{LN}_{l-1}(\mathbf{x}_l) + \mathcal{F}_l(\operatorname{LN}_{l}(\operatorname{LN}_{l-1}(\mathbf{x}_l))))\label{keel_as_post}
\end{equation}

Because the shortcut in \ours{} carries a normalized signal, it follows the structural definition of a \postln{} architecture.

\textbf{Non-Redundancy of Successive LayerNorms:}
A natural question arises with Equation~\ref{keel_as_post} - if the transformation branch $\mathcal{F}_l$ receives $\operatorname{LN}_{l}(\operatorname{LN}_{l-1}(\mathbf{x}_l))$, can these two normalization operations be merged into one? 

Mathematically, they cannot. Recall that a LayerNorm operation is defined as:
\begin{equation}
    \operatorname{LN}(\mathbf{x}) = \frac{\mathbf{x}}{\|\mathbf{x}\|_2} \odot \boldsymbol{\gamma},
\end{equation}
where $\gamma$ is learnable element-wise affine parameters. In \ours{}, the first normalization $\operatorname{LN}_{l-1}$ applies its own scaling $\gamma_{l-1}$ to the input. Because the subsequent $\operatorname{LN}_{l}$ re-calculates the normalization term \textit{after} this affine transformation has been applied, the two operations do not collapse into a single linear step. The motivation of designing two successive LayerNorms is discussed in Section~\ref{design_evol}.
\section{Discussions}
\label{sec:discussion}

\subsection{\model{} vs. DeepNorm}

The design of \ours{} draws inspiration from DeepNorm \cite{deepnorm}, a normalization approach originally proposed to stabilize deep Transformers. However, our empirical evidence suggests that while DeepNorm is effective for standard encoder-decoder architectures, its stability and performance degrade when applied to large-scale decoder-only LLMs, often underperforming standard \preln{} baselines. \ours{} addresses these shortcomings through two fundamental methodological divergences.

\textbf{Gradient Flow vs. Output Magnitude:}
The core theoretical premise of DeepNorm relies on bounding the forward output magnitude to prevent explosion. It derives scaling factors $\alpha$ specifically to ensure the variance of the output remains bounded as depth increases. However, bounding the forward pass does not inherently guarantee a healthy backward gradient flow. In contrast, \ours{} is derived explicitly from a gradient perspective (as detailed in Section \ref{sec:ana_keel}). By analyzing the Jacobian properties of the residual blocks, we identify that the primary instability in deep \postln{} models stems from vanishing gradients in the lower layers, not just exploding variance in the forward pass. Our formulation ensures that the gradient norm remains consistent ($\approx 1$) regardless of depth $L$, providing a more robust optimization signal for LLMs.

\textbf{Structural Topology vs. Initialization Dependence:}
DeepNorm relies heavily on a specialized initialization scheme (scaling weights by $L^{-0.25}$) to complement its architecture. This creates a strong dependency on the initial state of the network. In the context of LLMs, which undergo massive pre-training on trillions of tokens, the model weights drift significantly from their initialization. As the training dynamics evolve, the benefits of a specific initialization strategy often diminish, leading to instability in later stages of training. \ours{} eliminates this dependency by baking the stabilization mechanism into the architecture itself (via the scaling factor $\alpha$ and the auxiliary LN). This structural approach ensures stability throughout the entire training trajectory, rather than just the initial phase.

\subsection{Design Evolution: From Naive Scaling to \model{}}\label{design_evol}

To arrive at the optimal architecture, we conducted a step-by-step ablation study, incrementally refining the formulation based on gradient flow analysis.

\textbf{Attempt 1: Naive Residual Scaling.}
Our initial experiments adapted a modified DeepNorm \cite{deepnorm} approach to the \postln{} architecture. We scaled the shortcut connections by $\alpha = L$ without adopting DeepNorm's specific initialization constraints:
\begin{equation}
    \mathbf{x}_{l+1} = \operatorname{LN}\left(\alpha \mathbf{x}_{l} + \mathcal{F}_l(\mathbf{x}_{l})\right).
\end{equation}

\paragraph{Observation:} While this formulation exhibited better stability than the vanilla DeepNorm on LLMs, both its training stability and final convergence performance lagged behind the standard \preln{} baseline. We hypothesized that the unnormalized input to $\mathcal{F}_l$ resulted in high variance within the residual branch, complicating the optimization landscape.

\textbf{Attempt 2: Learnable Input Scaling.}
To control the variance entering the residual block, we introduced a learnable vector $\boldsymbol{\beta}$ (initialized to $\mathbf{1}$) to dynamically scale the input to $\mathcal{F}_l$:
\begin{equation}
    \mathbf{x}_{l+1} = \operatorname{LN}\left(\alpha \mathbf{x}_{l} + \mathcal{F}_l(\boldsymbol{\beta} \odot \mathbf{x}_{l})\right).
\end{equation}

\paragraph{Observation:} While this improved forward pass stability (as $\boldsymbol{\beta}$ learned to scale down inputs), it introduced a side effect: gradient attenuation. As $\boldsymbol{\beta}$ decreases to control variance, it proportionally scales down the gradients flowing back through $\mathcal{F}_l$, effectively choking the learning signal for the attention and FFN blocks. Consequently, stability remained inferior to \preln{}.

\textbf{Attempt 3: Decoupling Scale and Variance.}
To resolve the gradient attenuation, we sought to normalize the input distribution while maintaining learnable scaling. We introduced an explicit Layer Normalization step before the scaling factor $\boldsymbol{\beta}$:
\begin{equation}
    \mathbf{x}_{l+1} = \operatorname{LN}\left(\alpha \mathbf{x}_{l} + \mathcal{F}_l(\boldsymbol{\beta} \odot \operatorname{LN}(\mathbf{x}_{l}))\right).
\end{equation}
This addition is pivotal. As established in our theoretical analysis (Equation~\ref{eq:ffn}), the normalization ensures that the gradient magnitude through $\mathcal{F}$ is preserved:
\begin{equation} \label{eq:ffn}
    \frac{\partial \operatorname{LN}_{l,1}(\mathbf{z}_l)}{\partial \mathbf{z}_l}
    \frac{\partial \mathcal{F}}{\partial \operatorname{LN}_{l,2}( \mathbf{x}_{l})} 
    \frac{\partial \operatorname{LN}_{l,2}( \mathbf{x}_{l})}{\partial \mathbf{x}_{l}}
    = \mathcal{O}\left[ \frac{\gamma_{l,1}}{\sqrt{\gamma_{l-1,1}^2\alpha^2 + \gamma_{l,2}^2}} \left\| \frac{\partial \mathcal{F}}{\partial \operatorname{LN}_{l,2}( \mathbf{x}_{l})} \right\|_2 \frac{\gamma_{l,2}}{\gamma_{l-1,1}} \right].
\end{equation}
Crucially, the scaling factor $\gamma_{l,2}$ (associated with the inner LN) is effectively canceled by the affine weights $\gamma_{l-1,1}$ of the previous layer, preventing the gradient vanishing observed in Attempt 2.

\textbf{Final Formulation (\model{}).}
We observe that Equation (14) contains redundant parameters. The Layer Normalization operation $\operatorname{LN}(\mathbf{x})$ already includes a learnable affine weight vector $\boldsymbol{\gamma}$. Therefore, the external learnable vector $\boldsymbol{\beta}$ can be conceptually absorbed into the internal affine weights of the $\operatorname{LN}$. By merging these terms, we arrive at the concise and efficient formulation of \ours{}:
\begin{equation}
    \mathbf{x}_{l+1} = \operatorname{LN}\left(\alpha \mathbf{x}_{l} + \mathcal{F}_l(\operatorname{LN}(\mathbf{x}_{l}))\right).
\end{equation}

\subsection{Depth-wise Test-Time Training}

A prominent trend in modern architecture design involves replacing quadratic-complexity attention mechanisms with linear recurrent models. These linear attention paradigms can be formulated as a recurrent state update. For instance, standard linear attention updates its hidden state $\mathbf{S}_t$ as:
\begin{equation}
    \mathbf{S}_{t+1} = \mathbf{S}_{t} + \mathbf{v}_{t}\mathbf{k}_{t}^\top.
\end{equation}
Recent works \cite{deltanet, lact, titans} interpret this recurrence through the lens of \textit{Test-Time Training} (TTT). In this view, the state update is equivalent to a gradient descent step minimizing a transient objective function $\mathcal{L}_{\text{seq}}$ at each time step $t$:
\begin{equation} \label{ttt_seq}
    \mathcal{L}_{\text{seq}}^{(t)} = -\langle \mathbf{S}_t \mathbf{k}_t, \mathbf{v}_t \rangle.
\end{equation}
By maximizing the alignment between the retrieved state $\mathbf{S}_t \mathbf{k}_t$ and the value $\mathbf{v}_t$, the model "trains" itself on the context history.

We identify a structural isomorphism between this sequence-wise recurrence and the depth-wise propagation in Residual Networks. Consider the update rule for a standard \preln{} Transformer layer:
\begin{equation}
    \mathbf{x}_{l+1} = \mathbf{x}_{l} + \mathcal{G}(\mathbf{x}_{l}; \mathbf{W})\mathbf{W}_o^\top,
\end{equation}
where $\mathcal{G}(\cdot)$ represents the transformation block (Attention or FFN):
\begin{align}
    \mathcal{G}_{\text{attn}}(\mathbf{x}_{l}) &= \operatorname{Attention}(\mathbf{x}_l; \mathbf{W}_q, \mathbf{W}_k, \mathbf{W}_v), \\
    \mathcal{G}_{\text{FFN}}(\mathbf{x}_{l}) &= \sigma(\mathbf{W}_{in}\mathbf{x}_l).
\end{align}
This residual update can analogously be viewed as Test-Time Training along the \textit{depth} dimension. The layer output $\mathbf{x}_{l+1}$ is the result of a gradient step optimizing a depth-wise objective $\mathcal{L}_{\text{depth}}$:
\begin{equation} \label{ttt_depth}
    \mathcal{L}_{\text{depth}}^{(l)} = -\langle \mathbf{x}_l \mathbf{W}_o, \mathcal{G}(\mathbf{x}_l) \rangle.
\end{equation}
From this perspective, the architectural modifications in \ours{}, particularly the scaling factor $\alpha$ and the additional normalization, function as optimization prerequisites that stabilize the ``depth-wise training'' process. Furthermore, we observed that deep modeling exhibits phenomena similar to those in sequence modeling. For instance, in deep modeling, the first few layers are crucial, akin to the attention sink phenomenon~\cite{attn-sink, streaming-attn}. Moreover, both of them exhibit local patterns: in sequence modeling, recent tokens typically receive more attention, whereas in deep modeling, deeper layers become increasingly significant. Further details can be found in Appendix~\ref{ap:layer}.

This creates a compelling parallel with recent advancements in sequence modeling, such as Titans or LaCT. Just as these models introduce gating mechanisms (analogous to our $\alpha$) and state normalization (analogous to our residual LN) to improve the recurrent TTT objective in Equation~\ref{ttt_seq}, \ours{} applies similar principles to improve the depth-wise TTT objective in Equation~\ref{ttt_depth}. 

We believe this duality suggests a promising avenue for future research: techniques developed to improve long-context sequence recurrence can likely be adapted to develop infinite-depth model propagation, and vice versa.
\section{Experiments}

\subsection{Stability Analysis}

We evaluate the training stability of \ours{} against a comprehensive suite of established normalization strategies. Our baselines include the standard \postln{} and \preln{} architectures, as well as recent variants designed for stability: DeepNorm \cite{deepnorm}, and hybrid approaches such as HybridNorm \cite{hybridnorm} and Mix-LN \cite{mixln}.

\textbf{Benchmark Protocol: Maximum Tolerable Learning Rate.}
To quantify stability, we measure the \textit{Maximum Tolerable Learning Rate} (Max LR). It is defined as the highest learning rate a model can sustain during warm-up stage without diverging.

We adopt a stress-test protocol where the learning rate schedule is set with an aggressively high peak of $\eta_{\text{peak}} = 5 \times 10^{-2}$. The learning rate increases linearly from $0$ to $\eta_{\text{peak}}$ over the warm-up period. We monitor the training dynamics at every step; the learning rate recorded at the exact step immediately preceding divergence is reported as the Max LR. A higher Max LR indicates a more robust optimization landscape and superior training stability.

\begin{figure}[h]
    \centering
    \begin{subfigure}[b]{0.32\linewidth}
        \centering
        \includegraphics[width=\linewidth]{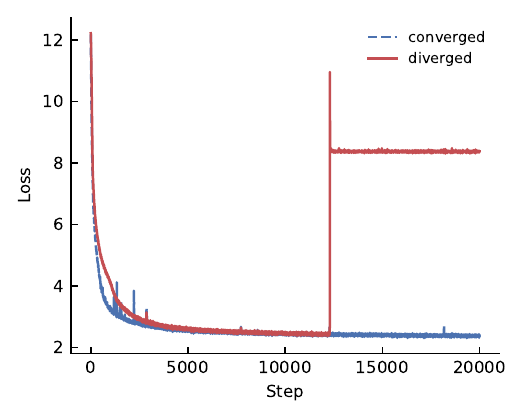}
        \caption{Loss stagnation}
        \label{fig:loss_stagnation}
    \end{subfigure}
    \begin{subfigure}[b]{0.32\linewidth}
        \centering
        \includegraphics[width=\linewidth]{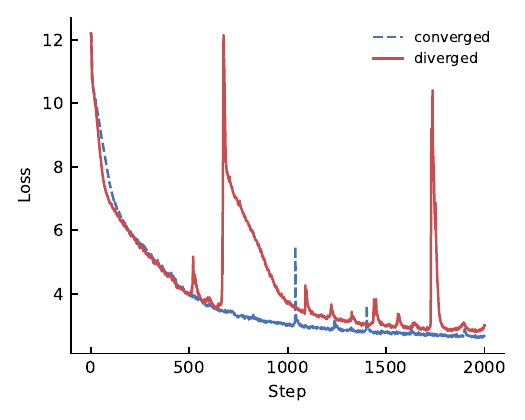}
        \caption{Irrecoverable instability}
        \label{fig:irrecoverable_instability}
    \end{subfigure}
    \begin{subfigure}[b]{0.32\linewidth}
        \centering
        \includegraphics[width=\linewidth]{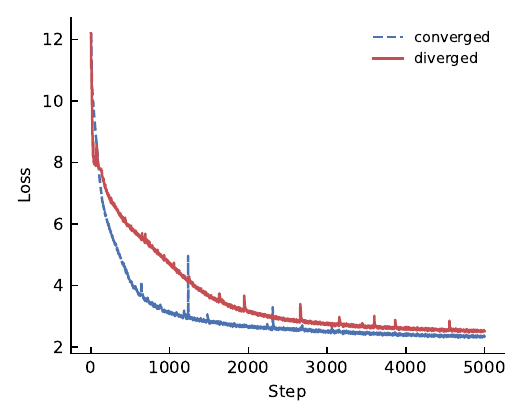}
        \caption{Optimization degradation}
        \label{fig:optimization_degradation}
    \end{subfigure}
    \caption{The illustration of three distinct pathological behaviors of model divergence in the training of LLMs.}
    \label{fig:criteria}
\end{figure}

\textbf{Criteria for Divergence.}
Identifying divergence in LLMs goes beyond simple numerical overflow (NaN). Based on our observations, we categorize divergence into three distinct pathological behaviors:

\begin{enumerate}
    \item \textit{Loss Stagnation:} The loss curve enters a plateau significantly earlier than expected, ceasing to decrease despite continued training. This often indicates vanishing gradients or saturation in the normalization layers.
    \item \textit{Irrecoverable Instability:} The loss exhibits high-magnitude spikes from which the model fails to recover. Unlike transient spikes common in LLM training, these result in a permanent degradation of the loss value (i.e., the loss never returns to the pre-spike baseline).
    \item \textit{Optimization Degradation:} The model does not exhibit explicit spikes, and the loss continues to decrease. However, the convergence rate is anomalously slow compared to a healthy baseline run. This ``silent'' failure mode typically suggests that the effective update step size has been excessively dampened.
\end{enumerate}

If a training run satisfies any of the above criteria, it is marked as diverged.

\textbf{Results.}
Table~\ref{tab:fineweb_lr} summarizes the stability limits across varying depths (64 and 512 layers). \ours{} demonstrates a significant improvement in stability compared to all baselines.

Notably, in the 64-layer configuration, \ours{} tolerates a learning rate of $1.01 \times 10^{-2}$, surpassing the standard \preln{} ($7.65 \times 10^{-3}$) and exceeding the vanilla \postln{} baseline by nearly two order of magnitude. This trend holds for extremely deep networks: at 512 layers, \ours{} maintains a Max LR of $6.31 \times 10^{-3}$, validating our theoretical claims regarding training stability in deep architectures.

\begin{table}[t]
    \centering
    \setlength{\tabcolsep}{5pt}
    \renewcommand{\arraystretch}{1.1}
    \small
    \begin{tabular}{l|cccccc}
    \toprule
     & \textbf{Post-LN} & \textbf{DeepNorm} & \textbf{HybridNorm} & \textbf{Mix-LN} &  \textbf{Pre-LN} & \textbf{\model{}} \\
    \midrule
    \multicolumn{7}{c}{\textit{Configuration A: 64 Layers, 5,000 Warm-ups, $\eta_{\text{peak}} = 5 \times 10^{-2}$}} \\
    \midrule
    \textbf{Max LR} & $3.0 \times 10^{-4}$ & $3.5 \times 10^{-4}$ & $4.9 \times 10^{-4}$ & $8.6 \times 10^{-4}$ & $7.65 \times 10^{-3}$ & $\mathbf{1.01 \times 10^{-2}}$ \\
    \midrule
    \multicolumn{7}{c}{\textit{Configuration B: 512 Layers, 5,000 Warm-ups, $\eta_{\text{peak}} = 5 \times 10^{-2}$}} \\
    \midrule
    \textbf{Max LR} & $2.8 \times 10^{-4}$ & $3.5 \times 10^{-4}$ & $3.5 \times 10^{-4}$ & $3.5 \times 10^{-4}$ & $4.67 \times 10^{-3}$ & $\mathbf{6.31 \times 10^{-3}}$ \\
    \bottomrule
    \end{tabular}
    \caption{Comparison of Maximum Tolerable Learning Rates across different architectures. \ours{} consistently supports higher learning rates, indicating superior training stability, particularly as model depth increases to 256 layers.}
    \label{tab:fineweb_lr}
\end{table}

\begin{figure}[t]
    \centering
    \begin{subfigure}[b]{0.45\linewidth}
        \centering
        \includegraphics[width=\linewidth]{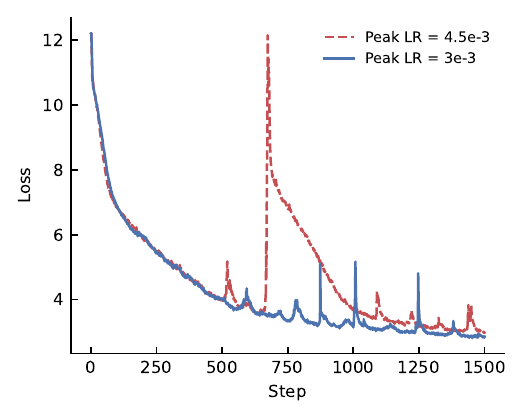}
        \subcaption{256 layers, batch size 8M}
        \label{fig:preln-loss-a}
    \end{subfigure}
    \begin{subfigure}[b]{0.45\linewidth}
        \centering
        \includegraphics[width=\linewidth]{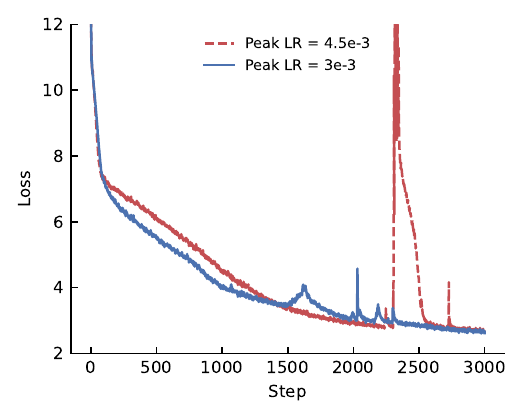}
        \subcaption{512 layers, batch size 4M}
        \label{fig:preln-loss-b}
    \end{subfigure}
    \caption{Training loss curves of \preln{} during the early stage of training. \preln{} exhibits a pronounced loss spike when trained with a higher learning rate.}
    \label{fig:preln-loss}
\end{figure}

\subsection{Optimal Learning Rate}

Having established the theoretical stability of \ours{}, we investigate its impact on downstream performance. We hypothesize that the ability to tolerate larger learning rates is not merely a stability metric, but an optimization advantage that allows the model to traverse the loss landscape more effectively and escape local minima.

To verify this, we conducted a controlled sweep of peak learning rates ($\eta_{\text{peak}} \in \{1.5, 3.0, 6.0\} \times 10^{-3}$) for both the \preln{} baseline and \ours{}. All models have 512 layers and a hidden dimension of 1024. We pre-train these models with the same 250B tokens from the internal data. The learning rate linearly increases from 0 to $\eta_{\text{peak}}$ over the first 2500 steps, then concludes with a cosine decay to $1.0 \times10^{-7}$. 

As summarized in Table~\ref{tab:lr_sweep}, the results reveal two critical optimization behaviors. The \preln{} baseline exhibits performance saturation and instability at higher learning rates. While some tasks (e.g., MMLU) improve at $\eta = 6.0\times10^{-3}$, others suffer from degradation (e.g., ARC-Easy drops from 75.8 to 74.0; MBPP drops from 25.0 to 22.8), suggesting that the model is oscillating in specific subspaces. 
In contrast, \ours{} demonstrates a consistent, monotonic performance gain across all benchmarks as the learning rate increases. At $\eta = 6.0\times10^{-3}$, \ours{} achieves a global average of 55.5, significantly outperforming the best Pre-LN configuration (52.3).

The benefits of stable, high-learning-rate training are most pronounced in reasoning-intensive tasks. On GSM-8K, increasing the learning rate to $6.0\times10^{-3}$ boosts \ours{} to 43.8, a massive improvement over the Pre-LN baseline (38.1). This confirms that \ours{} effectively unlocks the optimization potential of large step sizes without suffering from the gradient vanishing or instability typical of deep LLMs.

\begin{table}[t!]
    \centering
    \renewcommand{\arraystretch}{1.2}
    \setlength{\tabcolsep}{6.8pt}
    
    \begin{tabular}{l | ccc | ccc}
    \toprule
    \multicolumn{1}{c|}{\textbf{Benchmark}} & \multicolumn{3}{c|}{\textbf{Pre-LN}} & \multicolumn{3}{c}{\textbf{\model{}}} \\
    \cmidrule(lr){1-1} \cmidrule(lr){2-4} \cmidrule(lr){5-7}
    \multicolumn{1}{c|}{\textbf{Peak LR} \textit{\scriptsize ($\times 10^{-3}$)}} & \textbf{1.5} & \textbf{3.0} & \textbf{6.0} & \textbf{1.5} & \textbf{3.0} & \textbf{6.0} \\
    \midrule
    \multicolumn{7}{l}{\textit{\textbf{Common Sense \& Knowledge}}} \\
    \midrule
    MMLU {\scriptsize (5-shot)}         & 47.5 & 48.1 & 52.9 & 49.8 & 53.5 & \textbf{56.3} \\
    ARC-Easy {\scriptsize (25-shot)}    & 75.8 & 75.3 & 74.0 & 74.5 & 76.4 & \textbf{77.1} \\
    ARC-Challenge {\scriptsize (25-shot)}& 45.9 & 44.4 & 43.6 & 45.1 & 44.5 & \textbf{48.9} \\
    HellaSwag {\scriptsize (0-shot)}    & 64.2 & 64.2 & 64.9 & 64.3 & 65.5 & \textbf{67.4} \\
    LAMBADA {\scriptsize (0-shot)}      & 66.7 & 66.3 & 67.0 & 66.5 & 67.9 & \textbf{68.8} \\
    PIQA {\scriptsize (0-shot)}         & 75.6 & 75.5 & 75.1 & 76.3 & 75.8 & \textbf{76.7} \\
    AGI-Eval {\scriptsize (0-shot)}     & 29.3 & 29.7 & 34.7 & 30.5 & 34.6 & \textbf{39.6} \\
    Winogrande {\scriptsize (0-shot)}   & 64.1 & 64.2 & 63.5 & 63.8 & 64.5 & \textbf{65.7} \\
    CommonsenseQA {\scriptsize (0-shot)}& 48.6 & 46.6 & 55.7 & 53.2 & 52.3 & \textbf{61.3} \\
    
    \midrule
    \multicolumn{7}{l}{\textit{\textbf{Reasoning \& Coding}}} \\
    \midrule
    GSM-8K {\scriptsize (5-shot)}       & 30.3 & 31.1 & 38.1 & 30.5 & 36.6 & \textbf{43.8} \\
    HumanEval {\scriptsize (0-shot)}    & 14.0 & 17.7 & 17.7 & 14.6 & 18.3 & \textbf{19.5} \\
    MBPP {\scriptsize (0-shot)}         & 25.0 & 24.0 & 22.8 & 22.6 & \textbf{26.0} & \textbf{26.0} \\
    
    \midrule
    \multicolumn{7}{l}{\textit{\textbf{Multilingual Understanding}}} \\
    \midrule
    CMMLU {\scriptsize (5-shot)}        & 52.8 & 52.7 & 60.3 & 57.0 & 61.5 & \textbf{64.4} \\
    C-Eval {\scriptsize (5-shot)}       & 52.7 & 52.1 & 61.4 & 55.9 & 60.8 & \textbf{62.0} \\
    
    \midrule
    \textbf{Average Score} & 49.5 & 49.4 & 52.3 & 50.5 & 52.7 & \textbf{55.5} \\
    \bottomrule
    \end{tabular}
    \caption{\textbf{Performance scaling with Learning Rate.} While Pre-LN results are inconsistent at high learning rates (degrading on tasks like ARC-Easy and MBPP at $\eta=6.0\times10^{-3}$), \ours{} exhibits robust, monotonic improvement, effectively leveraging larger step sizes to achieve superior convergence.}
    \label{tab:lr_sweep}
\end{table}

\subsection{Scalability Analysis: Performance at Depth Scaling}

To validate the advantages of \ours{} in mitigating training instability, we conducted a comprehensive scaling study. We compared \ours{} against the robust \preln{} baseline across three distinct depth configurations: 64, 128, 512 and 1024 layers. All models underwent a rigorous two-stage training protocol consisting of general pre-training followed by continued pre-training (CPT). 

\textbf{Setup.} We use a batch size of 1024 and a sequence length of 4096. All models are first pre-trained on 190B tokens and then further pre-trained on an additional 60B tokens. We adopt the AdamW optimizer with $\beta_1=0.9$, $\beta_2=0.95$, and a weight decay of 0.01. Gradient clipping is applied with a maximum norm of 1.0. The learning rate schedule consists of a linear warmup over the first 2500 steps, followed by cosine decay to $1.0 \times10^{-7}$. We set the peak learning rate to $9.0\times 10^{-3}$ for the 64-layer and 128-layer models, and to $6.0\times 10^{-3}$ for the 512-layer models. For the 1024-layer models, we adopt a peak learning rate of $4.5\times 10^{-3}$ for \ours{} and $3.0\times 10^{-3}$ for \preln{}, as we observe that training 1024-layer \preln{} models with a peak learning rate of $4.5\times 10^{-3}$ leads to substantial instability (see Figure~\ref{fig:preln-loss-b}).

\textbf{Results.}
Table~\ref{tab:depth_scaling} details the zero-shot and few-shot performance across a diverse suite of benchmarks. The results reveal a clear trend: while \ours{} provides consistent gains at shallower depths (64L), its advantage becomes significantly more pronounced as the model depth increases, maintaining strong scalability even at 1024 layers.

\begin{itemize}
    \item \textbf{Reasoning Capabilities:} On complex reasoning tasks like GSM-8K and HumanEval, the performance gap widens dramatically at extreme depths. For instance, on GSM-8K, \ours{} achieves a score of 58.6 at 1024 layers, surpassing the 1024-layer \preln{} baseline (49.8) by 8.8 points. Notably, the \preln{} baseline shows signs of stagnation between 512 and 1024 layers on several reasoning tasks, whereas \ours{} continues to yield significant gains.
    \item \textbf{Global Average:} The average performance improvement scales significantly with depth. \ours{} improves over the baseline by approximately +1.7 points at 64 layers, +1.2 points at 128 layers, +3.8 points at 512 layers, and +3.0 points at 1024 layers.
\end{itemize}

This empirical evidence confirms that \ours{} effectively stabilizes optimization in ultra-deep LLMs, unlocking capabilities that are otherwise hindered by optimization difficulties or diminishing returns in standard architectures.

\begin{table}[t!]
    \centering
    \renewcommand{\arraystretch}{1.15}
    \setlength{\tabcolsep}{3.5pt}
    
    \begin{tabular}{l | cc | cc | cc | cc}
    \toprule
    \multirow{2}{*}{\textbf{Benchmark}} & \multicolumn{2}{c|}{\textbf{64 Layers}} & \multicolumn{2}{c|}{\textbf{128 Layers}} & \multicolumn{2}{c}{\textbf{512 Layers}} & \multicolumn{2}{c}{\textbf{1024 Layers}} \\
    \cmidrule(lr){2-3} \cmidrule(lr){4-5} \cmidrule(lr){6-7} \cmidrule(lr){8-9}
    & \textbf{Pre-LN} & \textbf{\model{}} & \textbf{Pre-LN} & \textbf{\model{}} & \textbf{Pre-LN} & \textbf{\model{}} & \textbf{Pre-LN} & \textbf{\model{}} \\
    
    \midrule
    \multicolumn{9}{l}{\textit{\textbf{Common Sense \& Knowledge}}} \\
    \midrule
    MMLU {\scriptsize (5-shot)} & 36.8 & \textbf{38.8} & 45.4 & \textbf{46.9} & 53.9 & \textbf{57.0} & 57.5 & \textbf{59.2} \\
    ARC-Easy {\scriptsize (25-shot)} & \bf 65.8 & 65.0 & \bf 69.7 & 68.7 & 76.6 & \textbf{78.6} & \textbf{80.4} & 80.3\\
    ARC-Challenge {\scriptsize (25-shot)} & \bf 33.7 & 33.6 & \bf 38.7 & 37.8 & 45.5 & \textbf{50.4} & 51.1 & \textbf{51.7}\\
    HellaSwag {\scriptsize (0-shot)} & 46.7 & \textbf{48.0} & 53.7 & \textbf{55.4} & 64.1 & \textbf{66.3} & 68.2 & \textbf{69.1}\\
    LAMBADA {\scriptsize (0-shot)} & 51.5 & \textbf{52.9} & 57.7 & \textbf{59.0} & 65.1 & \textbf{67.5} & 68.8 & \textbf{70.0}\\
    PIQA {\scriptsize (0-shot)} & \bf 69.3 & 69.0 & 71.7 & \textbf{72.1} & 75.0 & \textbf{76.3} & 77.0 & \textbf{78.3} \\
    AGI-Eval {\scriptsize (0-shot)} & 26.0 & \textbf{26.1} & 28.8 & \textbf{31.0} & 34.3 & \textbf{39.8} & 37.3 & \textbf{44.9}\\
    Winogrande {\scriptsize (0-shot)} & 57.4 & \textbf{58.2} & 59.5 & 59.5 & 64.2 & \textbf{65.4} & 66.1 & \textbf{67.0}\\
    CommonsenseQA {\scriptsize (0-shot)} & 31.4 & \textbf{37.8} & 48.2 & \textbf{49.1} & 60.0 & \textbf{65.8} & 60.6 & \textbf{67.1}\\
    
    \midrule
    \multicolumn{9}{l}{\textit{\textbf{Reasoning \& Coding}}} \\
    \midrule
    GSM-8K {\scriptsize (5-shot)} & 9.6 & \textbf{12.9} & 22.4 & \textbf{28.0} & 45.6 & \textbf{49.8} & 49.8 & \textbf{58.6}\\
    HumanEval {\scriptsize (0-shot)} & 9.8 & \textbf{11.0} & \bf 17.1 & 15.9 & 22.6 & \textbf{32.3} & 29.9 & \textbf{32.9}\\
    MBPP {\scriptsize (0-shot)} & 13.0 & \textbf{13.2} & 20.8 & \textbf{22.2} & 32.0 & \textbf{34.2} & 36.2 & \textbf{38.6}\\
    
    \midrule
    \multicolumn{7}{l}{\textit{\textbf{Multilingual Understanding}}} \\
    \midrule
    CMMLU {\scriptsize (5-shot)} & 39.3 & \textbf{43.4} & 50.1 & \textbf{52.7} & 61.0 & \textbf{65.0} & 64.9 & \textbf{67.0}\\
    C-Eval {\scriptsize (5-shot)} & 40.3 & \textbf{44.6} & 50.0 & \textbf{53.1} & 60.6 & \textbf{64.9} & 63.2 & \textbf{67.3}\\
    
    \midrule
    \textbf{Average Score} & 37.9 & \textbf{39.6} & 45.3 & \textbf{46.5} & 54.3 & \textbf{58.1} & 57.9 & \textbf{60.9}\\
    \bottomrule
    \end{tabular}
    \caption{\textbf{Scalability Benchmark.} Performance comparison between \preln{} and \ours{} across increasing model depths (64L, 128L, 512L, 1024L). The gain provided by \ours{} increases significantly with depth, particularly in reasoning-heavy tasks (GSM-8K, HumanEval), highlighting the method's effectiveness in stabilizing ultra-deep networks.}
    \label{tab:depth_scaling}
\end{table}

\subsection{Scalability Analysis: Performance at Data Scaling}

To assess the scalability of \ours{} with increasing training data, we compare it against Pre-LN across different token budgets. We train Pre-LN and \ours{} on 10B and 40B tokens from the FineWeb-EDU dataset~\cite{fineweb}. Both models have 256 layers and a hidden size of 1024, resulting up to 3B parameters. On the 10B-token run, we tune the peak learning rate for both \ours{} and the Pre-LN baseline over $\{1.0, 1.5, 3.0, 4.5\}\times 10^{-3}$. The batch size is set as 256. We use a maximum sequence length of 2048 for the 10B-token training and 4096 for the 40B-token training. The learning rate schedule uses a 2000-step linear warmup followed by cosine decay.

\begin{figure}[t]
    \centering
    \begin{subfigure}[b]{0.45\linewidth}
        \centering
        \includegraphics[width=\linewidth]{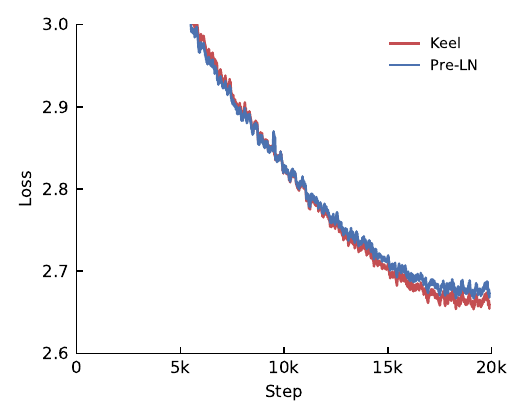}
        \subcaption{10B tokens, 512 layers}
        \label{fig:10b}
    \end{subfigure}
    \begin{subfigure}[b]{0.45\linewidth}
        \centering
        \includegraphics[width=\linewidth]{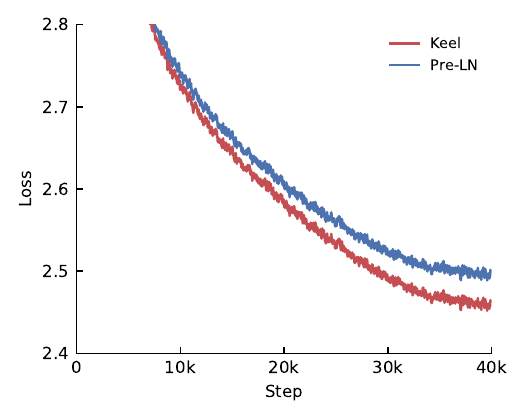}
        \subcaption{40B tokens, 512 layers}
        \label{fig:40b}
    \end{subfigure}
    \caption{Training loss curves of Pre-LN and \ours{} on FineWeb-EDU dataset~\cite{fineweb} with varying training tokens. As the training token scales from 10B to 40B tokens, \ours{} achieves larger gain on training loss compared to \preln{} baseline.}
    \label{fig:data}
\end{figure}

Figure~\ref{fig:data} presents the training loss curves of \ours{} and the baseline. As shown in Figure~\ref{fig:10b}, although \ours{} exhibits higher loss early in training, it overtakes Pre-LN as training proceeds and achieves lower loss by the end. More importantly, when scaling the training budget from 10B to 40B tokens, the performance gap further widens in favor of \ours{}. As shown in Table~\ref{tab:data-task}, the accuracy on various end tasks also exhibits similar trends: the improvement on HellaSwag increases from 0.9\% to 2.6\%. We attribute this to the lower effective depth of Pre-LN relative to \ours{}, which limits its representational capacity. As the amount of training data increases, the performance gain gradually plateaus. Overall, these results suggest that \ours{} is particularly well suited for large-scale training, while its advantages are less pronounced in low-data regimes.

\begin{table}[t]
    \centering
    \setlength{\tabcolsep}{5pt}
    \begin{tabular}{l|c|ccccccc}
    \toprule
    \textbf{Models}     & \makecell{\textbf{Best LR} \\ \scriptsize ($\times 10^{-3}$)}& \makecell{\textbf{ARC-Easy} \\ \small 25-shot} & \makecell{\textbf{PIQA} \\ \small 0-shot} & \makecell{\textbf{HellaSwag} \\ \small 0-shot} & \makecell{\textbf{LAMBADA} \\ \small 0-shot} & \makecell{\textbf{Winogrande} \\ \small 0-shot} & \makecell{\textbf{SciQ} \\ \small 0-shot} & \textbf{Average } \\
    \midrule
    \multicolumn{8}{l}{\textit{\textbf{10B tokens, 256 layers, 3B parameters}}} \\
    \midrule
    \preln{}     & 1.5 & 56.9 & 72.7 & 52.9 & \textbf{50.0} & 52.6 & 76.1 & 60.3\\
    \ours{}     & 3.0 & \textbf{58.8} & \textbf{73.4} & \textbf{53.8} & 49.6 & \textbf{56.0} & \textbf{78.1} & \textbf{61.5} \\
    \midrule
    \multicolumn{8}{l}{\textit{\textbf{40B tokens, 256 layers, 3B parameters}}} \\
    \midrule
    \preln{}     & 1.5 & \textbf{64.4} & 75.4 & 61.8 & 56.3 & 59.6 & 82.9 & 66.7 \\
    \ours{}     & 3.0 & 64.1  & \textbf{76.3} & \textbf{64.4} & \textbf{58.5} & \textbf{62.4} & \textbf{83.6} & \textbf{68.2}\\
    \bottomrule
    \end{tabular}
    \caption{Performance comparison between \preln{} and \ours{} across increasing training data (10B, 40B) on FineWeb-EDU dataset~\cite{fineweb}.}
    \label{tab:data-task}
\end{table}

\begin{table}[t]
    \centering
    \small
    \renewcommand{\arraystretch}{1.1}
    \setlength{\tabcolsep}{5pt}
    
    \begin{tabular}{l | c c | c }
    \toprule
    \textbf{Configuration} & \textbf{Deep (Pre-LN)} & \textbf{Wide (Pre-LN)} & \textbf{Deep (\model{})} \\
    \midrule
    \textit{Parameters} & 3B & 3B & 3B \\
    \textit{Layers} & 512 & 128 & 512 \\
    \textit{Hidden Dim} & 1024 & 2048 & 1024 \\
    \midrule
    \multicolumn{4}{l}{\textbf{Multilingual Understanding}} \\
    \midrule
    CMMLU {\scriptsize (5-shot)} & 60.3 & 59.3 & \textbf{64.4} \\
    C-Eval {\scriptsize (5-shot)} & 61.4 & 57.8 & \textbf{62.0} \\
    \midrule
    \multicolumn{4}{l}{\textbf{General Knowledge \& Commonsense}} \\
    \midrule
    MMLU {\scriptsize (5-shot)} & 52.9 & 51.5 & \textbf{56.3} \\
    ARC-Easy {\scriptsize (25-shot)} & 74.0 & 75.2 & \textbf{77.1} \\
    ARC-Challenge {\scriptsize (25-shot)} & 43.6 & 45.2 & \textbf{48.9} \\
    HellaSwag {\scriptsize (0-shot)} & 64.9 & 65.9 & \textbf{67.4} \\
    LAMBADA {\scriptsize (0-shot)} & 67.0 & 68.3 & \textbf{68.8} \\
    PIQA {\scriptsize (0-shot)} & 75.1 & 76.3 & \textbf{76.7} \\
    AGI-Eval {\scriptsize (0-shot)} & 34.7 & 36.1 & \textbf{39.6} \\
    Winogrande {\scriptsize (0-shot)} & 63.5 & 65.5 & \textbf{65.7} \\
    CommonsenseQA {\scriptsize (0-shot)} & 55.7 & 52.9 & \textbf{61.3} \\
    \midrule
    \multicolumn{4}{l}{\textbf{Math \& Code}} \\
    \midrule
    GSM-8K {\scriptsize (5-shot)} & 38.1 & 35.3 & \textbf{43.8} \\
    HumanEval {\scriptsize (0-shot)} & 17.7 & 16.5 & \textbf{19.5} \\
    MBPP {\scriptsize (0-shot)} & 22.8 & 24.4 & \textbf{26.0} \\
    \midrule
    \textbf{Average Score} & 52.3 & 52.2 & \textbf{55.5} \\
    \bottomrule
    \end{tabular}
    \caption{\textbf{Deeper vs. Wider.} Comparison of models with a fixed 3B parameter budget. While standard Deep models underperform Wide ones due to optimization difficulties, \ours{} effectively stabilizes the deep network, allowing it to outperform the Wide baseline significantly, particularly on reasoning-intensive tasks.}
    \label{tab:width_depth}
\end{table}

\subsection{Deeper vs. Wider}

A fundamental question in neural scaling laws is the optimal allocation of a fixed parameter budget: \emph{is it more effective to build deeper, narrower networks or shallower, wider ones?} Theoretically, deeper networks possess greater expressivity and reasoning depth. However, in practice, Wide topologies often outperform Deep ones because deep networks are notoriously difficult to train.

To investigate whether \ours{} overcomes this optimization barrier, we conducted a controlled experiment with a fixed parameter budget of 3B. We compare three configurations:
\begin{enumerate}
    \item \textbf{Baseline Deep (Pre-LN):} A standard deep topology (512 layers) which typically suffers from gradient degradation.
    \item \textbf{Baseline Wide (Pre-LN):} A standard wide topology (128 layers, 2048 hidden size), representing the industry standard for stability.
    \item \textbf{Ours (Deep):} The same deep topology (512 layers) augmented with \ours{}.
\end{enumerate}
We pre-train these models on 250B tokens of private data. We use a peak learning rate of $6.0\times10^{-3}$ for the 512-layer models and $3.0\times10^{-3}$ for the 128-layer models. We set the AdamW $\beta$ coefficients to (0.9, 0.95). The batch size and sequence length are set to 1024 and 4096, respectively. We use a warmup phase of 2500 steps.

\textbf{Results.}
Table~\ref{tab:width_depth} reports the results under a fixed budget of 3B parameters. We first compare the two \preln{} variants. Although the shallow-and-wide model achieves a lower training loss than the deep-and-narrow model (see Appendix~\ref{ap:mismatch}), their average downstream scores are close. This discrepancy suggests that training loss is not necessarily positively correlated with end-task performance, particularly when optimizing very deep LLMs. Meanwhile, the deep-and-narrow model exhibits a consistent advantage on more complex reasoning-oriented benchmarks (e.g., GSM-8K and MMLU), indicating that increasing depth has the potential to improve complex reasoning when the model can be trained effectively.

Motivated by these observations, \ours{} targets the stability and degradation issues of deep modeling. By stabilizing gradient flow and making deep training more reliable, \ours{} not only improves reasoning performance beyond the deep \preln{} baseline (e.g., 43.8 on GSM-8K vs.\ 38.1), but also achieves the best overall performance: our 512-layer \ours{} model reaches an average score of 55.5, outperforming both the deep \preln{} baseline (+3.2) and the wide \preln{} baseline (+3.3).

\subsection{Experimental Setup}

\begin{table}[t!]
    \centering
    \small
    \renewcommand{\arraystretch}{1.2}
    \setlength{\tabcolsep}{6pt}
    
    \begin{tabular}{l | c c}
    \toprule
    \textbf{Configuration} & \textbf{Pre-LN} & \textbf{\model{}} \\
    \midrule
    \textit{Architecture} & 512 Layers / 3B Params & 512 Layers / 3B Params \\
    \textit{Peak Learning Rate} & $3.0 \times 10^{-3}$ & $\mathbf{4.5 \times 10^{-3}}$ \\
    \midrule
    \multicolumn{3}{l}{\textit{\textbf{Multilingual Understanding}}} \\
    \midrule
    CMMLU {\scriptsize (5-shot)}        & 66.6 & \textbf{72.0} \\
    C-Eval {\scriptsize (5-shot)}       & 66.2 & \textbf{69.5} \\
    \midrule
    \multicolumn{3}{l}{\textit{\textbf{General Knowledge \& Commonsense}}} \\
    \midrule
    MMLU {\scriptsize (5-shot)}         & 59.5 & \textbf{62.7} \\
    ARC-Easy {\scriptsize (25-shot)}    & 79.7 & \textbf{81.6} \\
    ARC-Challenge {\scriptsize (25-shot)}& 51.6 & \textbf{53.6} \\
    HellaSwag {\scriptsize (0-shot)}    & 68.2 & \textbf{69.8} \\
    LAMBADA {\scriptsize (0-shot)}      & 68.0 & \textbf{69.7} \\
    PIQA {\scriptsize (0-shot)}         & 76.6 & \textbf{77.5} \\
    AGI-Eval {\scriptsize (0-shot)}     & 37.9 & \textbf{46.5} \\
    Winogrande {\scriptsize (0-shot)}   & 66.7 & 66.7 \\
    CommonsenseQA {\scriptsize (0-shot)}& 64.5 & \textbf{69.8} \\
    \midrule
    \multicolumn{3}{l}{\textit{\textbf{Math \& Code}}} \\
    \midrule
    GSM-8K {\scriptsize (5-shot)}       & 51.0 & \textbf{60.9} \\
    HumanEval {\scriptsize (0-shot)}    & 29.9 & \textbf{33.5} \\
    MBPP {\scriptsize (0-shot)}         & 35.0 & \textbf{40.6} \\
    \midrule
    \textbf{Average Score}              & 58.7 & \textbf{62.5} \\
    \bottomrule
    \end{tabular}
    \caption{\textbf{Pre-training Results (1T Tokens).} Comparison of a 512-layer \preln{} baseline vs. \ours{}. \ours{} capitalizes on its superior stability to train with a 50\% higher learning rate, resulting in substantial gains across all categories, particularly in reasoning-intensive tasks like GSM-8K and AGI-Eval.}
    \label{tab:main_pt}
\end{table}

To rigorously evaluate the scalability of our approach, we train a deep 512-layer LLM using the \ours{} formulation and compare it against a standard Pre-LN baseline. Both models utilize a hidden dimension of $d_{model}=1024$, resulting in an extreme depth-to-width ratio of $0.5$ (512 layers vs. 1024 width). This topology is specifically chosen to stress-test the optimization stability of deep networks where gradient signal preservation is critical. Both models contain approximately 3 Billion parameters.

Models are trained on a massive corpus of 1T tokens from our internal dataset. The training pipeline proceeds in two distinct phases: a general pre-training stage on the first 750B tokens, followed by continued pre-training (CPT) on the remaining 250B tokens to enhance reasoning and coding capabilities. The optimization is performed using AdamW with $\beta_1=0.9, \beta_2=0.95$, a weight decay of $0.01$, a global batch size of 2048, and a sequence length of 4096.

A critical differentiator in our setup is the learning rate schedule. We employ a linear warm-up over 2500 steps followed by cosine decay. \ours{} remains robust at higher learning rates, while the Pre-LN baseline becomes unstable, exhibiting pronounced loss spikes that force us to cap its peak rate at $3.0 \times 10^{-3}$ (see Figure~\ref{fig:preln-loss-a}). This stability allows us to train \ours{} at a significantly higher peak learning rate of $4.5 \times 10^{-3}$, unlocking superior convergence properties.

Following pre-training, both models undergo supervised fine-tuning (SFT) on a high-quality instruction mix. We perform a comprehensive grid search over learning rates ($\{1.0, 2.0, 3.0\} \times 10^{-6}$) and training durations (1 to 3 epochs) to ensure that the reported results reflect the optimal capability of each architecture. We utilize the lm-evaluation-harness~\cite{lm-eval} for standardized assessment across a broad suite of benchmarks spanning general knowledge~\cite{mmlu, arc, hellaswag, lambada, piqa, winogrande, commonsenseqa, agi-eval}, reasoning~\cite{gsm-8k}, coding~\cite{humaneval, mbpp}, and multilingual understanding~\cite{cmmlu, c-eval}.

\subsection{Main Results}

\textbf{Pre-training Performance.}
Table~\ref{tab:main_pt} summarizes the zero-shot and few-shot performance after the 1T token pre-training phase. \ours{} demonstrates a decisive advantage over the Pre-LN baseline, achieving a global average improvement of +3.8 points (62.5 vs. 58.7). A deeper analysis reveals that while \ours{} provides consistent gains across general knowledge tasks (e.g., MMLU, HellaSwag), the performance gap widens dramatically on tasks requiring complex, multi-step reasoning. For instance, on the GSM-8K math benchmark, \ours{} outperforms the baseline by nearly +10 points (60.9 vs. 51.0). Similarly, we observe significant improvements in code generation, with MBPP and HumanEval scores increasing by +5.6 and +3.6 respectively. This suggests that the improved gradient flow in \ours{} is particularly beneficial for learning the deep, hierarchical representations necessary for algorithmic reasoning, rather than simple pattern matching.

\textbf{Supervised Fine-Tuning (SFT) Results.}
We further investigate whether the advantages of \ours{} persist after supervised fine-tuning. Table~\ref{tab:sft_results} presents the results on a suite of challenging benchmarks, including MMLU-Pro and BBH (BIG-Bench Hard), which are designed to probe the limits of model capabilities. The results confirm that the ``pre-training advantage'' is effectively transferred to the SFT stage. The performance delta in reasoning tasks is preserved and, in some cases, amplified. On GSM-8K, the gap remains over 10 points (68.8 vs. 58.7), and on MMLU-Pro which requires nuanced understanding and robust instruction following, \ours{} achieves a score of 35.6 compared to the baseline's 26.6. This indicates that \ours{} not only learns better representations during pre-training but also possesses a more amenable optimization landscape for fine-tuning, allowing it to adapt to complex instructions without catastrophic forgetting or optimization difficulties.

\begin{table}[t!]
    \centering
    \small
    \renewcommand{\arraystretch}{1.2}
    \setlength{\tabcolsep}{8pt}
    
    \begin{tabular}{l | c c}
    \toprule
    \textbf{Benchmark} & \textbf{Pre-LN} & \textbf{\model{}} \\
    \midrule
    \multicolumn{3}{l}{\textit{\textbf{Reasoning \& Hard Tasks}}} \\
    \midrule
    MMLU-Pro {\scriptsize (5-shot)}     & 26.6 & \textbf{35.6} \\
    BBH {\scriptsize (3-shot)}          & 46.4 & \textbf{51.7} \\
    GSM-8K {\scriptsize (5-shot)}       & 58.7 & \textbf{68.8} \\
    HumanEval {\scriptsize (0-shot)}    & 31.7 & \textbf{37.2} \\
    MBPP {\scriptsize (0-shot)}         & 37.6 & \textbf{42.6} \\
    \midrule
    \multicolumn{3}{l}{\textit{\textbf{Knowledge \& Multilingual}}} \\
    \midrule
    MMLU {\scriptsize (5-shot)}         & 60.0 & \textbf{62.5} \\
    CMMLU {\scriptsize (5-shot)}        & 66.5 & \textbf{71.8} \\
    C-Eval {\scriptsize (5-shot)}       & 66.5 & \textbf{68.4} \\
    \midrule
    \textbf{Average Score}              & 49.3 & \textbf{54.8} \\
    \bottomrule
    \end{tabular}
    \caption{\textbf{SFT Performance.} After fine-tuning, \ours{} maintains its dominance, particularly on ``Hard'' benchmarks (BBH, MMLU-Pro) that test the limits of model reasoning, demonstrating that the architectural improvements translate directly to downstream tasks.}
    \label{tab:sft_results}
\end{table}

\section{Conclusion}

In this work, we propose that, as conventional scaling strategies plateau, depth is a promising axis for improving Transformer expressivity, which remained under-exploited due to optimization instability at extreme depths. We revisit Post-LayerNorm (Post-LN) formulation and identify its primary breakdown mechanism: the ResNet-style residual pathway induces gradient vanishing as depth grows, preventing reliable end-to-end signal propagation. To address this, we introduce \ours{}, a Post-LN Transformer that replaces the ResNet residual path with a Highway-style connection. This simple architectural change directly targets the source of vanishing by preserving gradient flow through the residual branch, enabling stable optimization without relying on specialized initialization schemes or elaborate training heuristics. Empirically, \ours{} trains robustly at depths exceeding 1000 sub-layers, and it delivers consistent improvements and more favorable depth-scaling behavior compared to widely used Pre-LN baselines. This reopens the design space for extreme-depth Transformers and points toward a practical path for exploring infinite-depth architectures in future work.

\section{Limitation and Future Work}
This work primarily focuses on how to stably train Post-LN LLMs when scaling depth. However, training instability is not driven by depth alone. As model width increases (e.g., hidden size, number of experts, or FFN dimension), LLM training can also become more unstable. In such wider settings, Post-LN may require a larger $\alpha$ (or stronger stabilization mechanisms) to maintain stable optimization. We leave a thorough investigation of LLM stability under width scaling to future work.

In addition, \ours{} addresses the effective-depth issue in Pre-LN LLMs. When the width-to-depth ratio is already high, this issue may be less pronounced, and the gains from \ours{} may therefore be less substantial. Finally, \ours{} typically requires substantial training data to be effective and is not recommended in low-data regimes.


\bibliographystyle{plainnat}
\bibliography{main}

\clearpage

\beginappendix

\section{Layer Redundancy in Deep LLMs}
\label{ap:layer}

\begin{figure}[h]
    \centering
    \includegraphics[width=\linewidth]{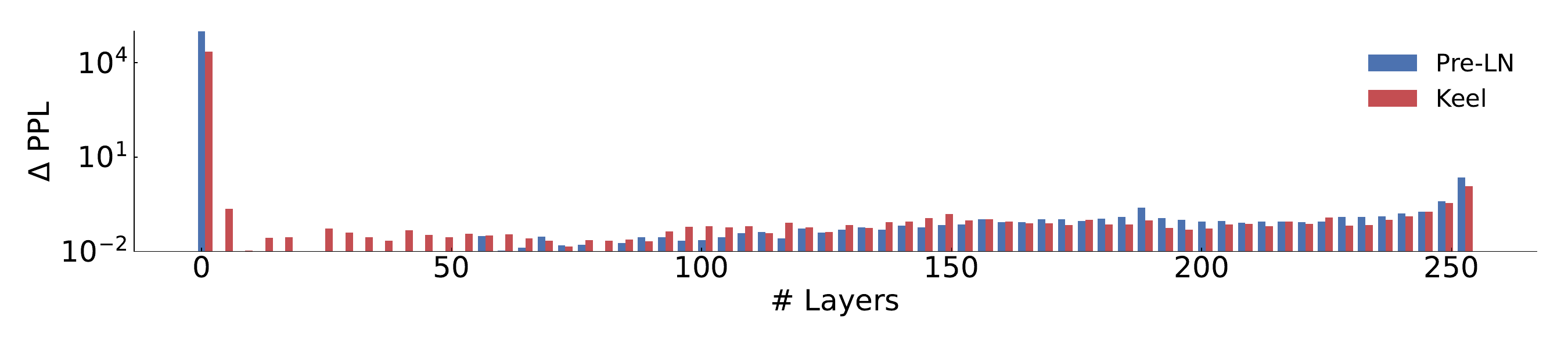}
    \caption{Performance degradation caused by the removal of each individual layer in \preln{} and \ours{}.}
    \label{fig:ppl_drop}
\end{figure}

\begin{figure}[h]
    \centering
    \begin{subfigure}[b]{\linewidth}
        \centering
        \includegraphics[width=\linewidth]{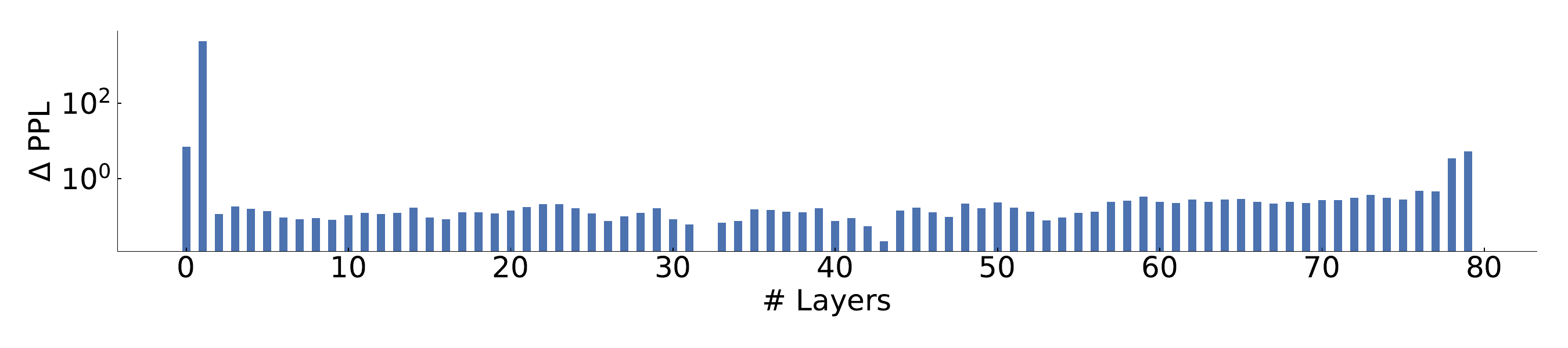}
        \caption{Qwen2.5-72B-Instruct}
        \label{fig:qwen-ppl-drop}
    \end{subfigure}
    \begin{subfigure}[b]{\linewidth}
        \centering
        \includegraphics[width=\linewidth]{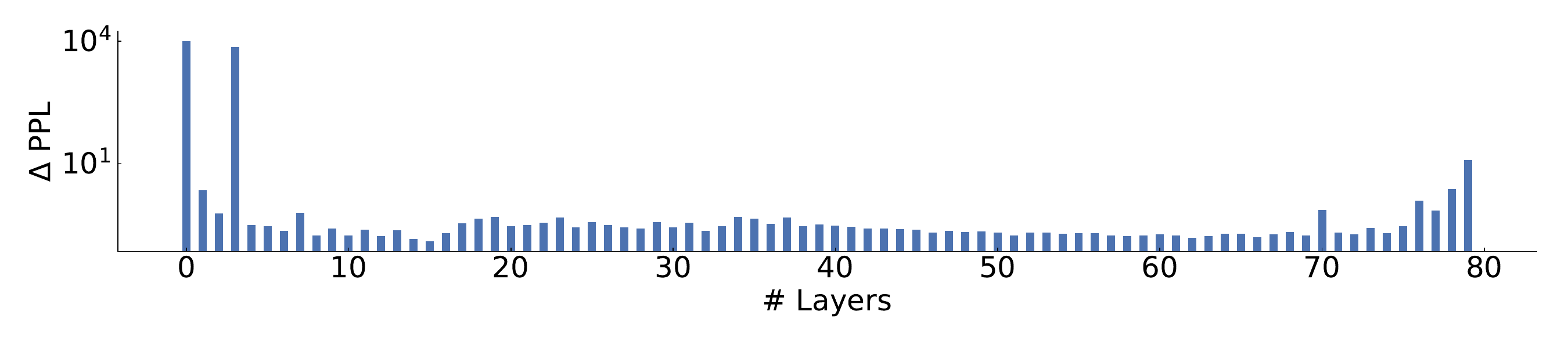}
        \caption{LLaMA-3.3-70B-Instruct}
        \label{fig:llama-ppl-drop}
    \end{subfigure}
    \caption{Performance degradation caused by the removal of each individual layer in Qwen2.5-72B-Instruct and LLaMA-3.3-70B-Instruct.}
    \label{fig:open-llm-drop}
\end{figure}

Layer redundancy constitutes a major challenge in training deep LLMs. Prior work~\cite{mixln, sun2025curse} suggests that layer redundancy is correlated with the placement of layer normalization. To quantify the redundancy of a specific layer in an LLM, we design a set of controlled ablation experiments. Concretely, we train the \preln{} and \ours{} on the same training data (250B tokens). We then measure layer redundancy by the performance degradation ($\Delta \text{PPL}_i = \text{PPL}_{\text{remove i-th layer}} - \text{PPL}_\text{full model}$) induced by removing an individual layer: a smaller drop indicates higher redundancy, while a larger drop indicates that the layer is more essential. For the open-source models, we select Qwen2.5-72B-Instruct~\cite{qwen2-5} and LLaMA-3.3-70B-Instruct~\cite{llama3}. Both models comprise 80 layers. We evaluate these models by computing perplexity on the C4 validation dataset~\cite{c4}. The results are demonstrated in Figure~\ref{fig:ppl_drop} and Figure~\ref{fig:open-llm-drop}.

We observe that \textbf{deep LLMs exhibit a phenomenon analogous to the “attention sink” reported in sequence modeling~\cite{attn-sink, streaming-attn}}. Specifically, the first few layers are exceptionally critical: removing it increases perplexity to above $10^4$, destroying model performance. Beyond the first layer, for both the Pre-LN model and the Post-LN-style Keel model, the performance degradation induced by layer removal increases with depth. We observe the same pattern for Qwen2.5-72B-Instruct and LLaMA-3.3-70B-Instruct. These results suggest that \textbf{shallower layers are more redundant, whereas deeper layers are increasingly pivotal}. Furthermore, compared to \preln{}, removing \ours{} layers produces a larger PPL increase in the shallow region, indicating that \ours{} substantially reduces shallow-layer redundancy and yields a greater effective depth.

\section{Discrepancy Between Training Loss and Downstream Evaluation}
\label{ap:mismatch}

Training loss is commonly used as a key metric for assessing the quality of LLM pre-training, as it is relatively simple to measure and often directly correlates with downstream task performance. As a result, it is frequently considered a critical indicator of model pre-training success. However, during the training of deep LLMs, we observe that a lower training loss does not necessarily correspond to better downstream performance.

\begin{figure}[h]
    \centering
    \begin{subfigure}[b]{0.45\linewidth}
        \centering
        \includegraphics[width=\linewidth]{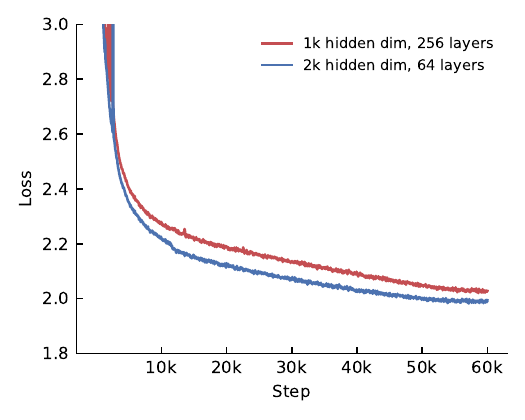}
        \caption{Training loss of shallow and deep Pre-LN.}
        \label{fig:mismatch-loss-a}
    \end{subfigure}
    \begin{subfigure}[b]{0.45\linewidth}
        \centering
        \includegraphics[width=\linewidth]{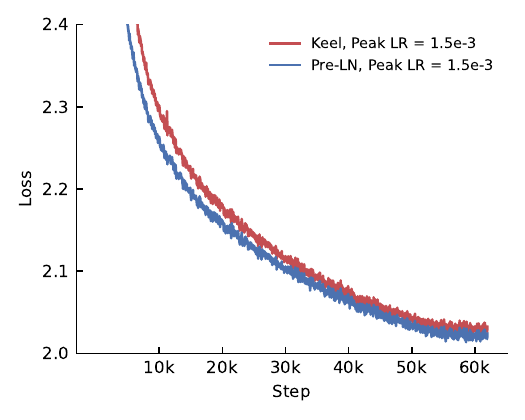}
        \caption{Training loss of Pre-LN and Keel.}
        \label{fig:mismatch-loss-b}
    \end{subfigure}
    \caption{Comparison of training loss in Pre-LN models and between Pre-LN and Keel models, illustrating the mismatch between training loss and downstream performance.}
    \label{fig:mismatch-loss}
\end{figure}

Figure~\ref{fig:mismatch-loss-a} shows the training loss curves of shallow and wide (2048 hidden size, 64 layers) and deep and narrow (1024 hidden size, 256 layers) Pre-LN models. Both models were trained on the same 250 billion tokens. While the shallow and wide Pre-LN model exhibits a significantly lower training loss, as shown in Table~\ref{tab:width_depth}, it performs slightly worse than the deep and narrow model on downstream tasks. 

A similar phenomenon was observed in the comparison between \preln{} and \ours{} models at a smaller learning rate. As shown in Figure~\ref{fig:mismatch-loss-b}, at a peak learning rate of $1.5\times10^{-3}$, the \preln{} model exhibits a slightly lower training loss than \ours{}, yet \ours{} demonstrates better performance on downstream tasks (see Table~\ref{tab:lr_sweep}). Above all, we find that in the training of deep LLMs, training loss and end-task performance are not always positively correlated. Therefore, evaluating end-task performance during pre-training is essential. We aim to further investigate the mechanisms behind this phenomenon in future work.

\section{Model Configuration}

\begin{table}[h]
    \centering
    \begin{tabular}{l|cc}
    \toprule
    \textbf{Configuration}     &  \multicolumn{2}{c}{\textbf{Hyper-parameters}} \\
    \midrule
    Hidden dimension     & 1024 & 2048 \\
    Intermediate dimension & 3072 & 6144 \\
    Attention heads & 16 & 16\\
    KV heads & 8 & 8 \\
    Initializer & $\mathcal{N}(0, 0.02^2)$ & $\mathcal{N}(0, 0.02^2)$ \\
    Dropout & 0.0 & 0.0 \\
    Attention dropout & 0.0 & 0.0 \\
    RMS Norm $\epsilon$ & 1e-5 & 1e-5 \\
    RoPE $\theta$ & 10000.0 & 10000.0 \\
    \bottomrule
    \end{tabular}
    \caption{Model configuration of \ours{} and Pre-LN baselines.}
    \label{tab:arch}
\end{table}

\end{document}